\documentclass[a4paper,fleqn]{cas-dc}

\usepackage[numbers]{natbib}

\usepackage{xcolor}
\usepackage{romannum}

\def\tsc#1{\csdef{#1}{\textsc{\lowercase{#1}}\xspace}}
\tsc{WGM}
\tsc{QE}
\tsc{EP}
\tsc{PMS}
\tsc{BEC}
\tsc{DE}
\begin{document}
\let\WriteBookmarks\relax
\def\floatpagepagefraction{1}
\def\textpagefraction{.001}
\shorttitle{Multimodal Spatio-Temporal Deep Learning Approach for Neonatal Postoperative Pain Assessment}
\shortauthors{M. S. Salekin et~al.}

\pagenumbering{arabic} % as romannum package changes this config

%%%%%%%%%%%%%%%%%%%%%%%%%%%%%%%%%%%%%%%%%%%%%%%%%%%%%%%%%%%%%%%%%%%%%%%%%%%%%%%%%%%%

\title [mode = title]{Multimodal Spatio-Temporal Deep Learning Approach for Neonatal Postoperative Pain Assessment}

\author[1]{{Md Sirajus} {Salekin}} 
% \cormark[1]
% \ead{salekin@usf.edu}

\author[1]{Ghada Zamzmi}
\author[1]{Dmitry Goldgof}
\author[1]{Rangachar Kasturi}
\author[2]{Thao Ho}
\author[1]{Yu Sun}

\address[1]{Department of Computer Science and Engineering, University of South Florida, Tampa, FL, USA}
\address[2]{College of Medicine Pediatrics, USF Health, University of South Florida, Tampa, FL, USA}

% \cortext[cor1]{Corresponding author. E-mail: salekin@usf.edu}
\nonumnote{* \copyright 2020. This manuscript version is made available under the CC-BY-NC-ND 4.0 license 
\href{http://creativecommons.org/licenses/by-nc-nd/4.0/}{http://creativecommons.org/licenses/by-nc-nd/4.0/}
}

%%%%%%%%%%%%%%%%%%%%%%%%%%%%%%%%%%%%%%%%%%%%%%%%%%%%%%%%%%%%%%%%%%%%%%%%%%%%%%%%%%%%

\begin{abstract} % 250 words
The current practice for assessing neonatal postoperative pain relies on bedside caregivers. This practice is subjective, inconsistent, slow, and discontinuous. To develop a reliable medical interpretation, several automated approaches have been proposed to enhance the current practice. These approaches are unimodal and focus mainly on assessing neonatal procedural (acute) pain. As pain is a multimodal emotion that is often expressed through multiple modalities, the multimodal assessment of pain is necessary especially in case of postoperative (acute prolonged) pain. Additionally, spatio-temporal analysis is more stable over time and has been proven to be highly effective at minimizing misclassification errors. In this paper, we present a novel multimodal spatio-temporal approach that integrates visual and vocal signals and uses them for assessing neonatal postoperative pain. We conduct comprehensive experiments to investigate the effectiveness of the proposed approach. We compare the performance of the multimodal and unimodal postoperative pain assessment, and measure the impact of temporal information integration. The experimental results, on a real-world dataset, show that the proposed multimodal spatio-temporal approach achieves the highest AUC (0.87) and accuracy (79\%), which are on average 6.67\% and 6.33\% higher than unimodal approaches. The results also show that the integration of temporal information markedly improves the performance as compared to the non-temporal approach as it captures changes in the pain dynamic. These results demonstrate that the proposed approach can be used as a viable alternative to manual assessment, which would tread a path toward fully automated pain monitoring in clinical settings, point-of-care testing, and homes.
\end{abstract}


\maketitle
%%%%%%%%%%%%%%%%%%%%%%%%%%%%%%%%%%%%%%%%%%%%%%%%%%%%%%%%%%%%%%%%

\section{Introduction}
Postoperative pain \cite{tiippana2016new} affects a large number of patients across the world, with an estimated number of 234 million surgical procedures each year \cite{weiser2015excess}. In case of neonates, more than 1.5 million anesthetics are performed every year in the United States for surgical procedures such as gastrostomy tube placement and circumcision \cite{weiser2015excess,dowell2017contribution}. This leads to the publications of a large body of research articles and guidelines in recent years to discuss optimal approaches for assessing and managing postoperative pain \cite{mcnair2004postoperative, taylor2006assessing, brasher2014postoperative, dmytriiev2019assessment}. Despite this significant attention, the management of postoperative pain has remained inadequate \cite{zamzmi2018review, quinlan2020postoperative, kulshrestha2019management}. This poor management is the main cause of delayed hospital discharge, which leads to substantial emotional and financial burden \cite{gan2017poorly, grosse2017employer}. In addition, it has been found \cite{gan2017poorly} that the poor management of postoperative pain can lead to serious short-term complications and long-term physiological, behavioral, and cognitive sequelae \cite{grunau2006long, walker2014neonatal}. As accurate pain assessment is the cornerstone for adequate management \cite{fortier2016pain}, it is critical to develop accurate pain assessment tools to obtain optimal interventions.

Broadly, pain in neonates can be categorized into three types \cite{stevens2007assessment}: acute procedural, acute prolonged, and chronic. Usually, prolonged acute pain (aka., postoperative pain) occurs after a major surgery (i.e. omphalocele repair), lasts for a longer time compared to acute procedural, and repeats with a decreasing rate after the surgery. The current practice \cite{walter2010pain} for assessing neonatal pain after a major surgery is manual and requires caregivers to observe specific behavioral (e.g., facial expression and body movement) and physiological (e.g., heart rate) indicators. Each of these indicators is assigned a score and the total pain score is generated by summing all the scores together. There are at least 29 validated score-based tools \cite{melo2014pain} for manually assessing procedural and postoperative pain in neonates, and more than half of these scales are multidimensional. The multidimensional pain assessment is necessary because pain manifests itself in various behavioral and physiological signals. Several studies (e.g., Ref. \cite{talbot2019sensory}) reported that pain has at least two dimensions, and suggested the use of multidimensional scales for effective assessment.

\begin{figure}
\centering
\includegraphics[width=.40\textwidth]{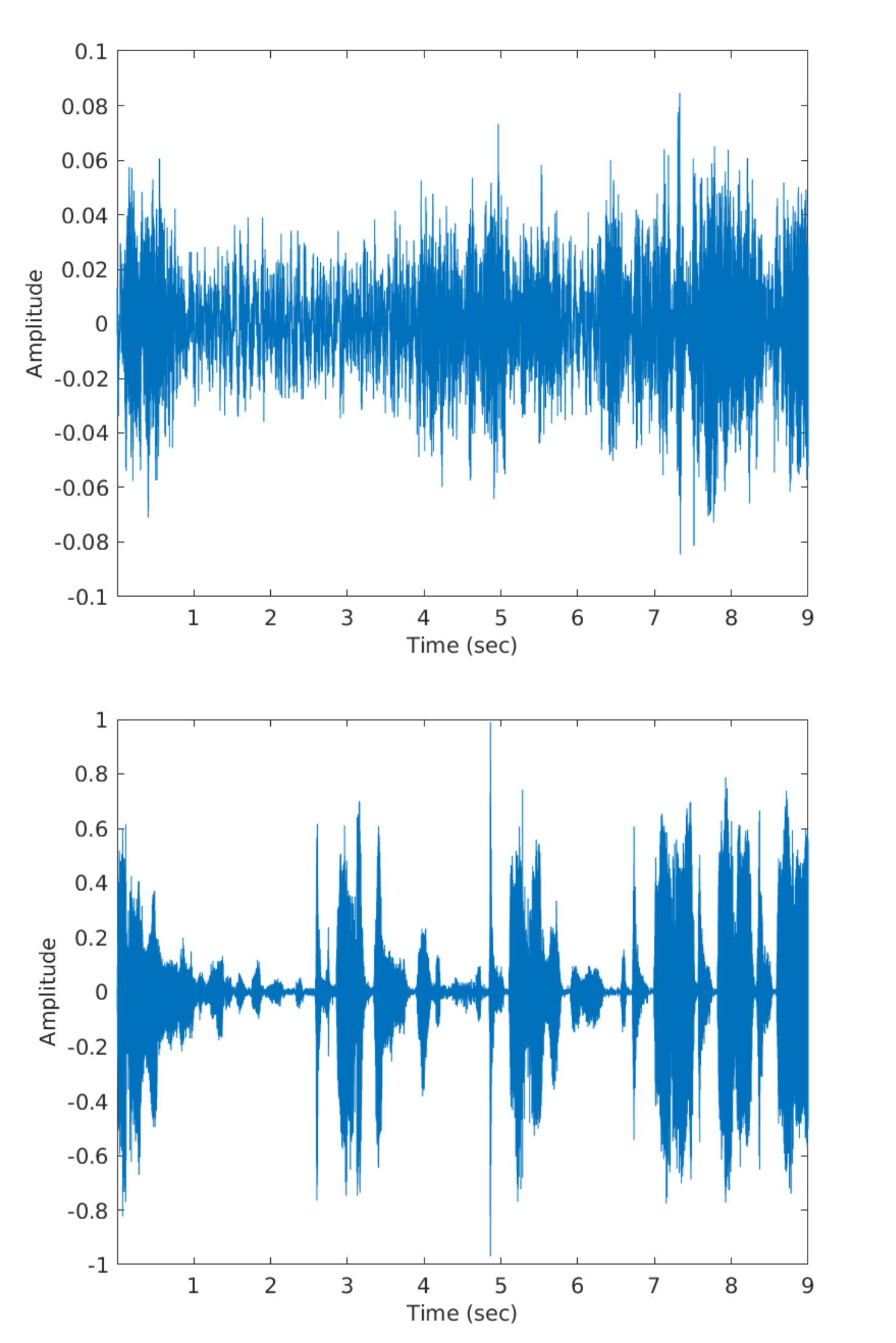}
\caption{Audio signals from procedural (top) and postoperative (bottom) pain. In both cases, the pain score of crying is 2. [Sample Rate = 44.1 kHz]. }
\label{Fig:procVpostop_audio}
\end{figure}

\begin{figure}
\centering
\includegraphics[width=.40\textwidth]{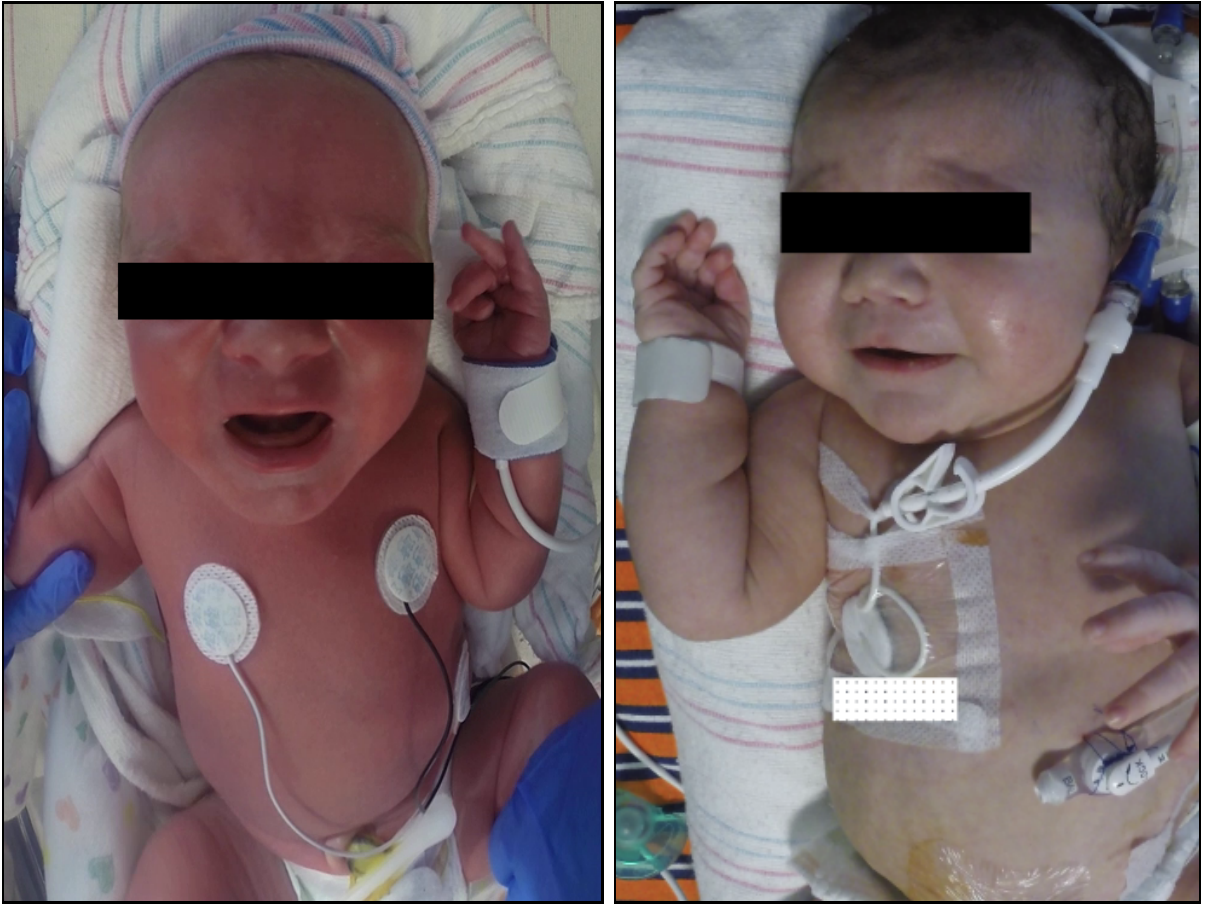}
\caption{Examples from neonatal procedural (left) and postoperative (right) pain. In both cases, the score of facial expression is 1.}
\label{Fig:procVpostop_visual}
\end{figure}

In addition, the multidimensional approach for assessment allows to 1) detect pain during the failure of recording a specific pain indicator due to developmental (e.g., facial nerve palsy), clinical (e.g., sedation), and environmental (e.g., background noise) factors, and 2) capture individual differences in pain reactions. The score-based multidimensional scales of procedural pain have a narrower range of scores (pain vs no-pain) as this type of pain tends to be intense for a short period of time and disappears as soon as its cause (e.g., heel lancing) is gone. On the contrary, acute prolonged (postoperative pain), or pain after any major surgery, continues long after its cause is gone, tends to have fluctuations in pain intensity, and evolves in a more complex pattern over time. Fig. \ref{Fig:procVpostop_audio} and Fig. \ref{Fig:procVpostop_visual} present examples of crying sounds and facial expression captured during procedural and postoperative pain, respectively. As can be seen, postoperative pain is less intense and occurs at different time intervals as compared to procedural pain (e.g., heel lancing). Hence, we believe assessing postoperative pain frequently and consistently is critical to generate effective plans for interventions.

The current practice for pain assessment using multidimensional score-based scales is discontinuous, inconsistent, and suffers from high inter- and intra-observer variations. To mitigate these limitations, several artificial intelligence-based methods \cite{sikka2015automated, celona2017neonatal, zamzmi2019comprehensive, seok2019postoperative} have been published in the literature. Of these published works, very few focus on assessing postoperative pain. For example, in Ref. \cite{sikka2015automated}, an automated method for assessing children postoperative pain based on the analysis of facial expression was proposed. Specifically, the proposed method extracts facial features around different facial action units (AUs) using handcrafted descriptors. The extracted features are then used to train a Support Vector Machine (SVM) to detect different levels of pain. Recently, a deep learning-based approach was proposed in Ref. \cite{salekin2020first} to assess the postoperative pain of neonates based on the analysis of facial expression. Instead of using a single indicator (unimodal) approach for the automated analysis of pain, in Ref. \cite{zamzmi2019comprehensive}, a multimodal approach that integrates facial expression, body movement, crying sound, and vital signs to assess procedural (short-term) pain of neonates has been proposed. The proposed approach used different handcrafted descriptors to extract pain-relevant features followed by training machine learning classifiers and fusing the output of these classifiers to obtain the pain label. Other works that propose automated methods for assessing neonatal procedural, or short-term, pain can be found in Ref. \cite{zamzmi2018review, celona2017neonatal, salekin2019multi, seok2019postoperative}. 

To summarize, the majority of existing machine learning approaches for pain assessment focus on procedural pain, are unimodal and do not take into account temporal information and dynamic pattern of pain. A recent multimodal approach \cite{zamzmi2019comprehensive}  was proposed to assess procedural acute pain using handcrafted methods, but it does not integrate temporal information. In this work, we propose the first spatio-temporal and multimodal AI-based approach for assessing neonatal postoperative pain. The main contributions of this paper can be summarized as follows.
\begin{itemize}
    \item We propose a novel temporal multimodal deep learning approach to assess neonatal postoperative pain. Existing works focus on assessing procedural, or short-term, pain based on the spatial analysis of a single pain indicator or traditional approaches using multiple pain indicators.
    \item We investigate and compare the performance of unimodal and multimodal approaches for assessing neonatal postoperative pain. We also compare the performance of the proposed multimodal approach with the state-of-the-art.  
    \item We present a multimodal pain dataset that includes video, audio, and physiological signals recorded from neonates during their hospitalization in the Neonatal Intensive Care Unit (NICU). The dataset is recorded during the normal state (baseline) as well as procedural (short-term) and postoperative (long-term) pain states.
\end{itemize}

The rest of the paper is organized as follows. Section \ref{Sec:technical_background} presents technical background needed to understand the rest of the paper. Section \ref{Sec:neonatal_pain_dataset} presents the neonatal pain dataset. Our approach is presented in Section \ref{Sec:methodology} followed by the experimental results in Section \ref{Sec:results_discussion}. Finally, Section \ref{Sec:conclusion} concludes the paper and discusses directions for future research.

\section{Technical Background}
\label{Sec:technical_background}
\subsection{VGG-Net and LSTM}
VGG-Net \cite{simonyan2014very} is a state-of-the-art Convolutional Neural Networks (CNN) for visual feature extraction. Although several versions of VGG-Net exist, VGG-16 \cite{simonyan2014very} has been widely and successfully used \cite{parkhi2015deep, cao2018vggface2}. VGG-16 \cite{simonyan2014very} consists of 13 uniform convolution layers followed by 3 fully connected layers. Each convolution layer uses a $3 \times 3$ kernel-size filters and is followed by a pooling layer. The network starts with 64 depth and gradually increases by a factor of 2 until it reaches 512. The depth of the network and the use of small kernel size allow to extract robust visual features. In this paper, we used VGG-16 \cite{simonyan2014very} network to extract visual features from the face, body, and spectrogram images of sounds. 

Long Short Term Memory (LSTM) \cite{hochreiter1997long} is one type of Recurrent Current Neural Networks (RNN) that is capable of learning the temporal information in a given sequence. Although RNN can handle long-term dependencies in theory, these networks fail to learn these dependencies in practice. To solve this issue, LSTM \cite{hochreiter1997long} network was introduced and has been widely used in a wide range of applications. LSTM \cite{hochreiter1997long} solves the long-term dependencies as well as vanishing gradient problem using the cell state, which is controlled by three gates: input, forget, and output gates. The input gate controls which information should be saved to the cell state. The forget gate controls which information should be ignored or forgotten from the previous cell state. Finally, the output gate controls which information should be sent to the next state. In this paper, we used LSTM \cite{hochreiter1997long} with the deep features, extracted by VGG-Net, to learn the temporal pattern and dynamics of postoperative pain.

\subsection{Bilinear CNN} 
Bilinear CNN \cite{lin2015bilinear} is introduced to address fine-grained image classification. It uses two CNN streams to extract features from two different regions of the same image, and the final bilinear vector is generated by combining the features of the two CNN streams. Mathematically, given that there are two CNN stream $X$ and $Y$ with pooling layer $P$ and classification layer $C$, then the bilinear model can be represented as $B = (X, Y, P, C)$. Now for a location $L$ within the image $I$, if the feature functions are $F_X$ and $F_Y$, then the bilinear feature vector $b$, can be represented as follows.
\begin{equation}
 b = (I, L, F_X, F_Y) \xrightarrow{} F_X(I, L)^T F_Y(I, L)
\end{equation}
Finally, a sum-pooling is applied to collect all the bilinear features from the entire image. To improve the performance, the final bilinear vector $u = \sum b{(I, L)}$ is forwarded to the following steps.
\begin{equation}
    v \xleftarrow[]{ sqrt}(sign(u)*\sqrt{\left|{u}\right|})
\end{equation}
\begin{equation}
    w \xleftarrow[]{normalization}(v/{||{v}||_2})
\end{equation}
The bilinear feature vector extracts orderless features, which provide better texture representation as compared to the order-full features in the fine-grained image classification problem. As discussed in Ref. \cite{lin2015bilinear}, this network is capable of extracting robust features in the context of the different pose, lighting, and background \cite{salekin2020first}. This resembles the context of the real-world NICU environment. In this paper, we used two VGG-16 \cite{simonyan2014very} models as CNN streams of the Bilinear CNN.

\section{Neonatal Pain Dataset}
\label{Sec:neonatal_pain_dataset}
To evaluate our temporal multimodal approach, we used a dataset containing data of procedural (acute) and postoperative (acute prolonged) neonatal pain. The dataset, which is known as USF-MNPAD-\Romannum{1} (University of South Florida Multimodal Neonatal Pain Assessment Dataset), was collected at the NICU in Tampa General Hospital, FL, USA. The dataset consists of 45 neonates with a gestational age that ranges from 30 to 41 weeks. It has ethnically and racially diverse population including Asian, African American, and Caucasian neonates. The data collection was approved by the USF Ethics Review Board (IRB \# Pro00014318).

\subsection{Setup and Painful Procedures}
USF-MNPAD-\Romannum{1} dataset has video, audio, and physiological data. To collect the video and audio data, a GoPro Hero Black 5 camera was used. The camera was set up on a camera stand facing the infant's incubator to capture the neonate's face and body. A bedside vital sign Phillips MP-70 monitor was used to collect the physiological data including heart rate, blood pressure, and oxygen saturation. All these data were recorded from neonates experiencing either short-term procedural or postoperative pan during their NICU hospitalization. The dataset contains multimodal data for 36 neonates (17 female) recorded during baseline, during a procedural pain stimulus (i.e., heel lancing and immunization), and immediately after the completion of the stimulus. In case of postoperative pain, 9 neonates (5 males) were recorded prior to major surgery (e.g., omphalocele-repair) to get their baseline state and monitored for three hours after the surgery to get their postoperative pain state. Note that in the current dataset, we only monitored the neonates up to three hours after the surgery due to clinical constraints.

\subsection{Ground Truth Labels}
The ground truth labels for both types of pain were documented independently by trained nurses using NIPS (Neonatal Infant Pain Scale) \cite{hudson2002validation} and N-PASS (Neonatal Pain, Agitation and Sedation Scale) \cite{hummel2008clinical} for procedural and postoperative pain, respectively. NIPS \cite{hudson2002validation} score-based pain scale has a total pain score that ranges from 0 to 7, and three levels of pain: no-pain (total score of 0-2), moderate pain (total score of 3-4), and severe pain (total score $>$ 4). The final score is generated by summing the individual scores of the following pain indicators: facial expression (score of 0 or 1), crying sound (score of 0, 1, or 2), breathing patterns (score of 0 or 1), arms movement (score of 0 or 1), legs movement (score of 0 or 1), and state of arousal (score of 0 or 1). 

N-PASS \cite{hummel2008clinical} score-based pain scale has a total score that ranges from -10 to +10, and five levels: deep sedation (score -10 to -5), light sedation (score -5 to -2), normal (score 0-2), moderate pain (score 3-5), and severe pain (score $>$ 5). This total score is generated by summing the individual scores of the following pain indicators: crying irritability, behavior state, facial expression, extremities of tone, and vital signs (heart rate, blood pressure, oxygen saturation). Each of these indicators has a score that ranges from -2 to +2, where minus (-), 0, and plus (+) indicate the sedation, normal, and pain states, respectively. In our dataset, we have 109, 33, and 76 samples for the normal state, moderate pain, and severe pain, respectively.

\begin{figure}
\centering
\includegraphics[width=.48\textwidth]{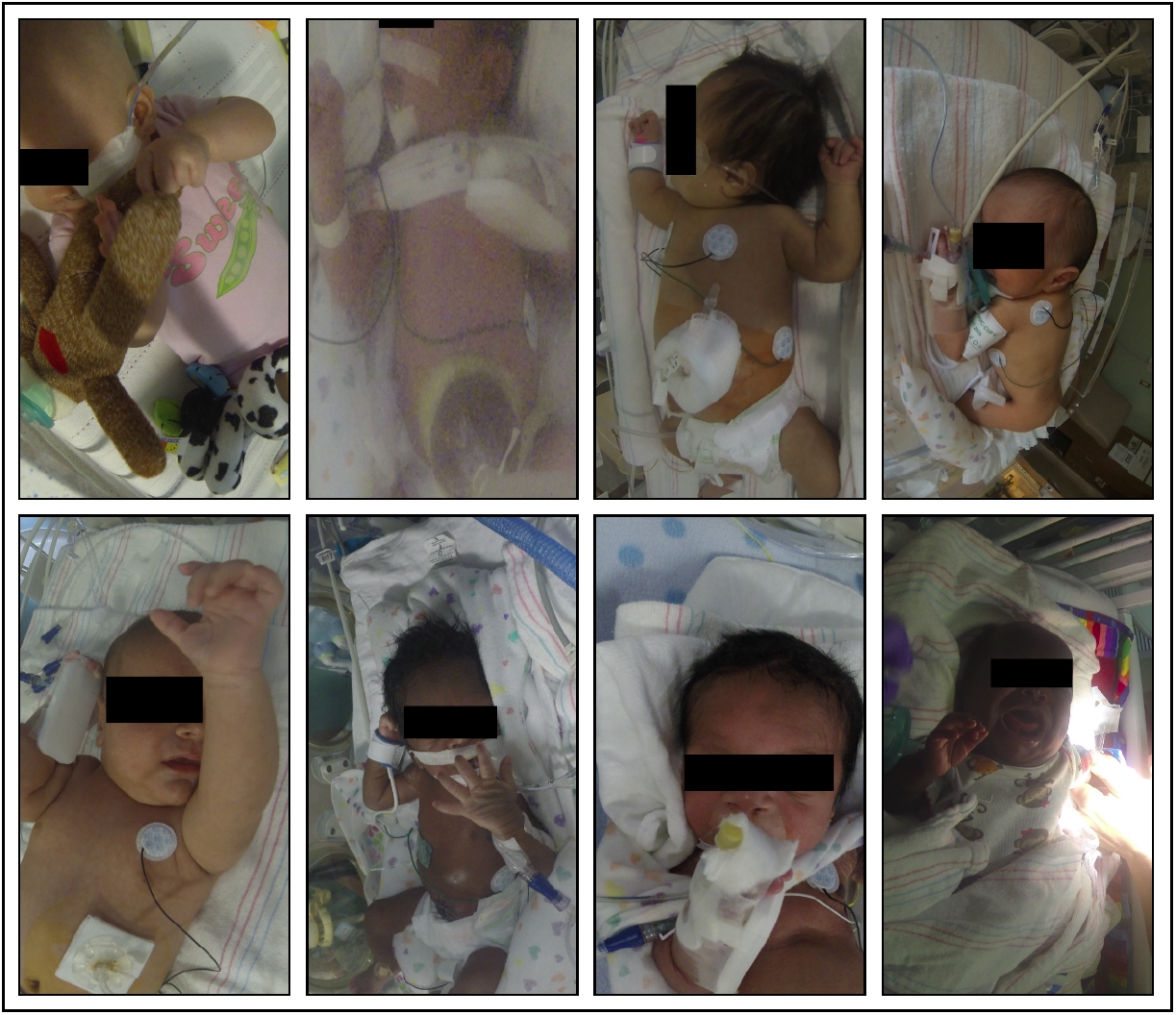}
\caption{Examples from our real-world neonatal postoperative dataset.}
\label{Fig:data_sample}
\end{figure}

Our dataset was labeled manually by independent trained nurses. The agreement between the nurses is measured using Kappa coefficient (0.85) and Pearson correlation (0.89). We include all the cases of agreement and exclude the cases of disagreement from further analysis. Fig. \ref{Fig:data_sample} shows examples from neonates recorded during postoperative pain. The images were randomly selected and masked to ensure confidentiality. 

\section{Methodology}
\label{Sec:methodology}
\begin{figure*}
\centering
\includegraphics[width=.95\textwidth]{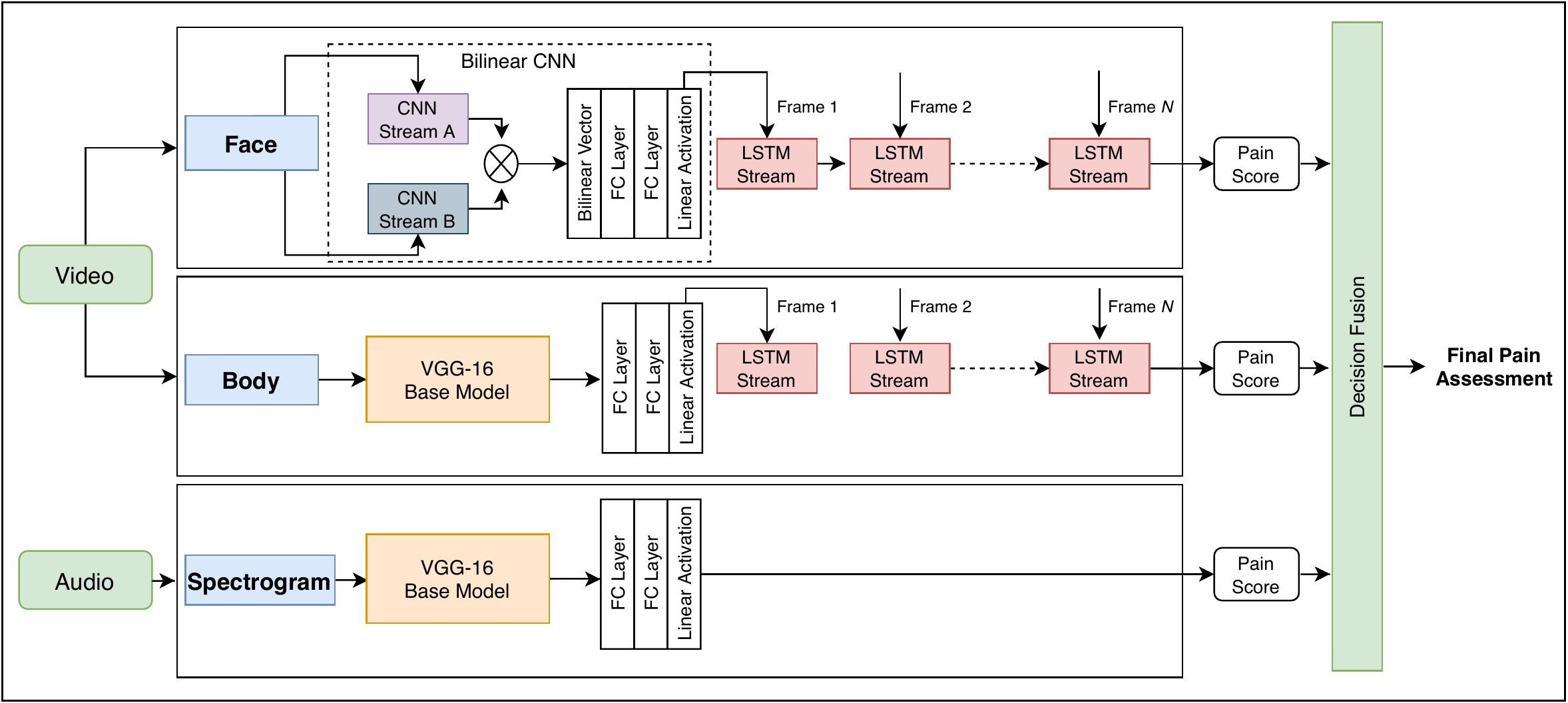}
\caption{Flowchart of the proposed spatio-temporal multimodal approach for neonatal postoperative pain assessment.}
\label{Fig:multimodal_approach}
\end{figure*}

In this paper, we investigate the use of a temporal multimodal approach for assessing postoperative pain. Our approach combines facial expression, body movement, and crying sound. We used the data of procedural and postoperative pain (see Section \ref{Sec:neonatal_pain_dataset}) for separately training different models corresponding to different pain indicators. For each pain indicator, spatio-temporal features are extracted and used to generate the score of that specific indicator. Then, we fused the scores of all indicators to generate the final pain level. Fig. \ref{Fig:multimodal_approach} represents an overview of the proposed temporal multimodal approach for assessing postoperative pain. 

\subsection{Facial Expression Analysis}
\subsubsection{Pre-processing and Augmentation}\label{Sec:methodology_1.1}
The first pre-processing step involves extracting key-frames from all videos using the FFmpeg library\footnote{https://ffmpeg.org/}. We then detected the face region in each frame using a pre-trained YOLO-based \cite{redmon2018yolov3} face detector. The YOLO face detector was pre-trained using the WIDER face dataset \cite{yang2016wider}, which contains around 393,703 faces. We empirically decided to fix the total number of key-frames extracted from each video segment to 32 frames. Using a fixed number of frames is important because the number of key-frames in each video varies. Further, the face region in some key-frames was occluded, which causes the face detector to fail. Therefore, we used a fixed number of key-frames to facilitate the training process. We randomly dropped some key-frames if the number of frames is larger than 32 and used resampling techniques to generate more frames if the number is lower than 32. To enlarge the dataset prior to the CNNs training, we performed image augmentation on the key-frames using random composition of 30$^{\circ}$ rotation, $\pm 25\%$ brightness change, and horizontal flipping.

\subsubsection{Facial Feature Extraction}
Deep learning-based architectures (e.g., VGG-Net) have been successfully used for detecting a wide range of emotions including pain \cite{rodriguez2017deep, celona2017neonatal, haque2018deep, zamzmi2019convolutional, salekin2020first}. In this paper, we fine-tuned a pre-trained VGG-16 \cite{simonyan2014very} CNN architecture to extract visual features from images captured during postoperative pain. Table \ref{Tab:vgg16_architecture} shows the details of the fine-tuned VGG-16 \cite{simonyan2014very} architecture. Since empirical evidence showed that Bilinear CNN (Section \ref{Sec:technical_background}) can better capture subtle changes, we used a Bilinear CNN with two VGG-16 \cite{simonyan2014very} streams to learn pain-related features. As shown in Fig. \ref{Fig:multimodal_approach}, the features extracted by both streams are then combined to generate the bilinear vector followed by two Fully Connected (FC) layers (64 units) and a dense layer (1 unit, $linear$ activation). Also, Dropout layers (0.5) are added after each FC layers to prevent over-fitting. We used two VGG-16 networks, which were pre-trained using VGGFace2 \cite{cao2018vggface2} and ImageNet \cite{deng2009imagenet} datasets, as the streams of the Bilinear CNN. We then fine-tuned the entire Bilinear CNN model using our procedural and postoperative dataset.

% VGG-16 configuration
\begin{table}
\caption{Details of fine-tuned VGG-16 architecture.} 
\begin{tabular*}{\tblwidth}{@{} LL@{} }
   \toprule
    Layer Type & Configuration\\
    \midrule
    Base model & Before FC layer without Pooling \\
    \hline
    FC & Dense  512, Relu\\
    \hline
    Dropout & Dropout (0.5) \\
    \hline
    FC & Dense 512, Relu\\
    \hline
    Dropout & Dropout (0.5) \\
    \hline
    FC & Dense 1, Activation = Linear \\
    \bottomrule
\end{tabular*}

\label{Tab:vgg16_architecture}
\end{table}

\subsubsection{Temporal Information Integration}
Pain is a dynamic event that evolves in a particular pattern over time. Hence, it is necessary to integrate temporal information to obtain an accurate assessment of pain \cite{rodriguez2017deep, haque2018deep, salekin2019multi}. After extracting the features using the Bilinear CNN, the deep features are further trained by RNN to learn the pain dynamics. Specifically, we used LSTM \cite{hochreiter1997long} network with the configuration shown in Table \ref{Tab:rnn_architecture}. We used two LSTM layers followed by two FC layers. Finally, a Dense layer with $sigmoid$ activation was used to classify the signal as pain or no-pain. To prevent over-fitting, dropout layers were used as shown in Table \ref{Tab:rnn_architecture}.

% RNN configuration
\begin{table}
\caption{Details of LSTM architecture.} 
\begin{tabular*}{\tblwidth}{@{} LL@{} }
    \toprule
    Layer Type & Configuration\\
    \midrule
    
    \multirow{3}{*}{RNN} & LSTM 16, Activation = Tanh, \\
   & Recurrent Activation = Hard Sigmoid, \\ & Dropout (0.2) \\

    \hline
    \multirow{3}{*}{RNN} & LSTM 16, Activation = Tanh, \\
                         & Recurrent Activation = Hard Sigmoid, \\ & Dropout (0.2) \\
    \hline
    FC & Dense  16, Relu\\
    \hline
    Dropout & Dropout (0.3) \\
    \hline
    FC & Dense 16, Relu\\
    \hline 
    Dropout & Dropout (0.3) \\
    \hline
    FC & Dense 1, Activation = Sigmoid\\
    \bottomrule
\end{tabular*}
\label{Tab:rnn_architecture}
\end{table}

\subsection{Body Movement Analysis}

\subsubsection{Pre-processing and Augmentation}
Similar to the facial expression (Section \ref{Sec:methodology_1.1}), we extracted the key-frames from the video segments using FFmpeg library. We used a YOLO detector, which was pre-trained originally on COCO dataset \cite{lin2014microsoft} containing around 330K images from 80 object categories, to detect the body regions of neonates. Further, similar to facial expression, we fixed the number of key-frames to 32 from each video segment. The resampling technique helps us to generate an equal number of frames in case of any failure detection. To enlarge the dataset for the CNN training, we performed random composition of rotation (30$^{\circ}$), brightness change ($\pm 25\%$), and horizontal flipping.

\subsubsection{Feature Extraction}
The state-of-the-art methods for extracting pain-relevant features from body regions are handcrafted-based (e.g., motion image) and deep-learning-based (e.g., VGG-16 \cite{simonyan2014very}). Therefore, we used two types of method, namely the motion image and VGG-16 \cite{simonyan2014very}, to assess neonatal postoperative pain from body movement.

\begin{figure}
\centering
\includegraphics[width=.48\textwidth]{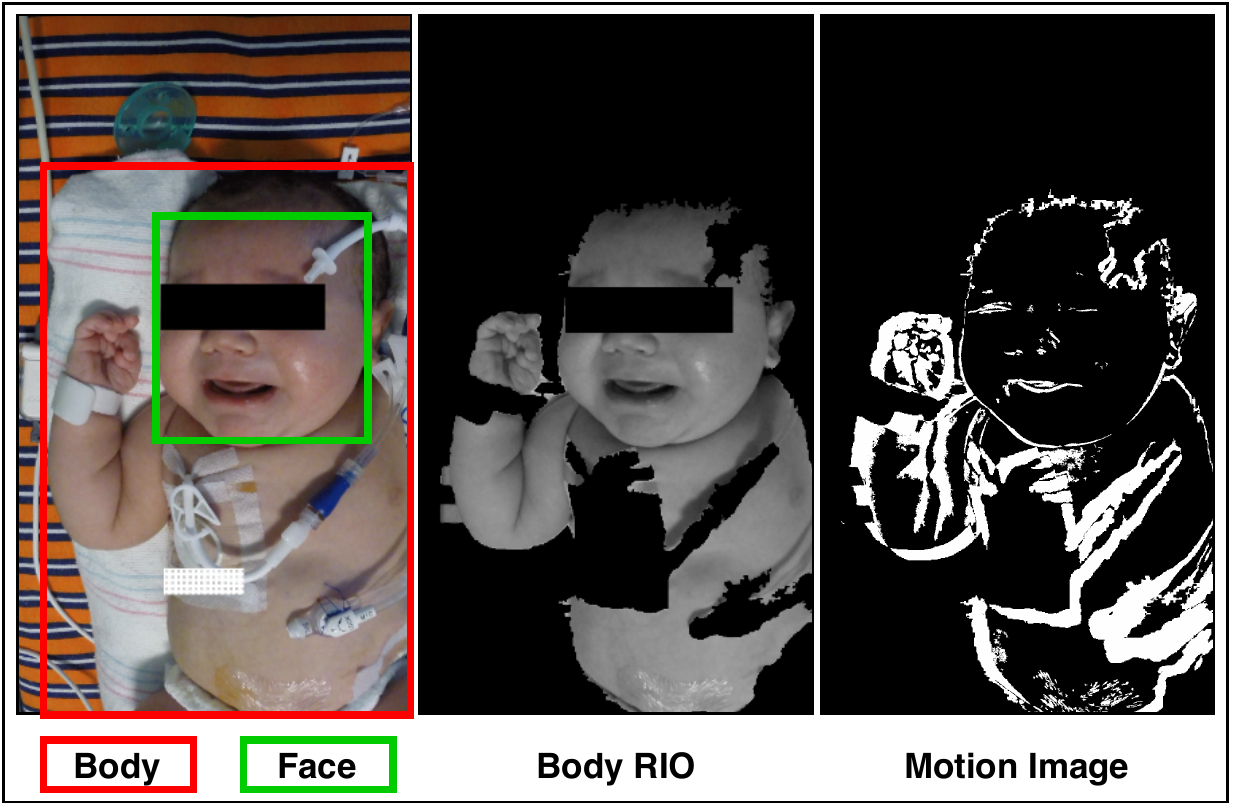}
\caption{Region of Interest (RIO) from sample input image.}
\label{Fig:input_roi}
\end{figure}

The motion image identifies the changes in pixels between consecutive frames, and it is calculated by subtracting consecutive frames followed by thresholding. Pixels of the motion image have a value of 1 (movement) and 0 (no-movement). To calculate the total motion in each frame, all the pixels are summed together and divided by the frame's dimensions. The calculated total motion is then used as the main feature \cite{zamzmi2019comprehensive} to train traditional classifiers such as Gaussian Naive Bayes \cite{rish2001empirical}, Random Forest \cite{breiman2001random}, and K-Nearest Neighbors \cite{altman1992introduction}. For deep learning, we trained the VGG-16 \cite{simonyan2014very} networks using both the motion image and original body image. The configurations of the fine-tuned VGG-16 \cite{simonyan2014very} network are presented in Table \ref{Tab:vgg16_architecture}. Fig. \ref{Fig:input_roi} shows different ROIs (Region of Interest) of a sample subject.

\subsubsection{Temporal Information Integration}
To capture the temporal changes of body movement, we integrated RNN (i.e. LSTM \cite{hochreiter1997long}) network to VGG-16 \cite{simonyan2014very}. We used the same LSTM \cite{hochreiter1997long} network architecture (Table \ref{Tab:rnn_architecture}), which is used for the facial expression (see Table \ref{Tab:rnn_architecture}). The integration of VGG-16 \cite{simonyan2014very} and LSTM \cite{hochreiter1997long} allows to learn body movement dynamics over time.

\subsection{Crying Sound Analysis}

\subsubsection{Pre-processing and Augmentation}
During the failure of recording a specific pain indicator due to occlusion or swaddle, crying sound can be used to assess pain. The state-of-the-art methods for extracting pain-relevant features from crying sounds are handcrafted-based (e.g., MFCC \cite{zamzmi2019comprehensive}) and deep-learning-based (e.g., spectrogram image \cite{salekin2019harnessing}). Therefore, we extracted two types of features, MFCC, and deep features, and used them to assess neonatal postoperative pain. 

MFCC, which stands for Mel Frequency Cepstral Coefficient (MFCC), is a popular Cepstral Domain \cite{oppenheim2004frequency} method that has been successfully used to extract a useful and representative set of features (i.e., coefficients) from an audio signal while discarding noise and non-useful features. Taking the Inverse Fourier Transform (IFT) of the logarithm of the signal's spectrum converts the audio signal to the Cepstral Domain. We extracted 20 MFCCs features over all of the frames of an audio segment (approx. 9 seconds). We then calculated the mean features from the 20 MFCCs, which lead to a mean MFCCs feature vector length of 388. 

In addition to MFCCs features, we converted the raw audio signal (approx.  9 seconds) to a spectrogram image. The spectrogram image \cite{oppenheim1999discrete} shows the visual representation of a given audio signal. It represents the change of frequency components with respect to time and suppresses noise. Brighter pixels in the spectrogram image represent higher energy and vice versa. After generating the spectrogram image for each audio segment, we extracted deep features from these images using a VGG-16 network. 

\begin{figure}
\centering
\includegraphics[width=.48\textwidth]{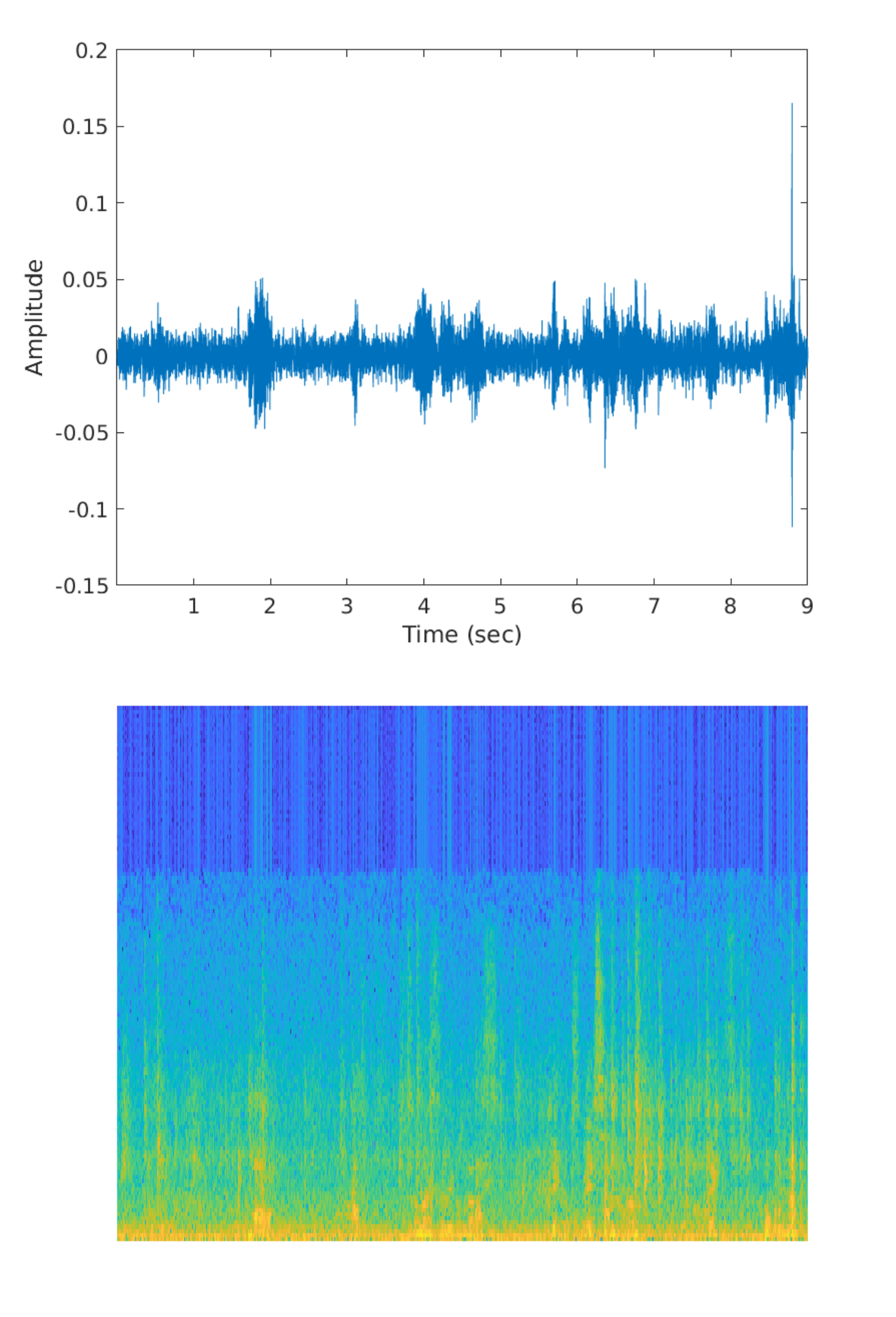}
\caption{Audio signal (top) and its corresponding spectrogram image (bottom) of a neonate during no-pain state. }
\label{Fig:spectrogram_nopain}
\end{figure}

\begin{figure}
\centering
\includegraphics[width=.48\textwidth]{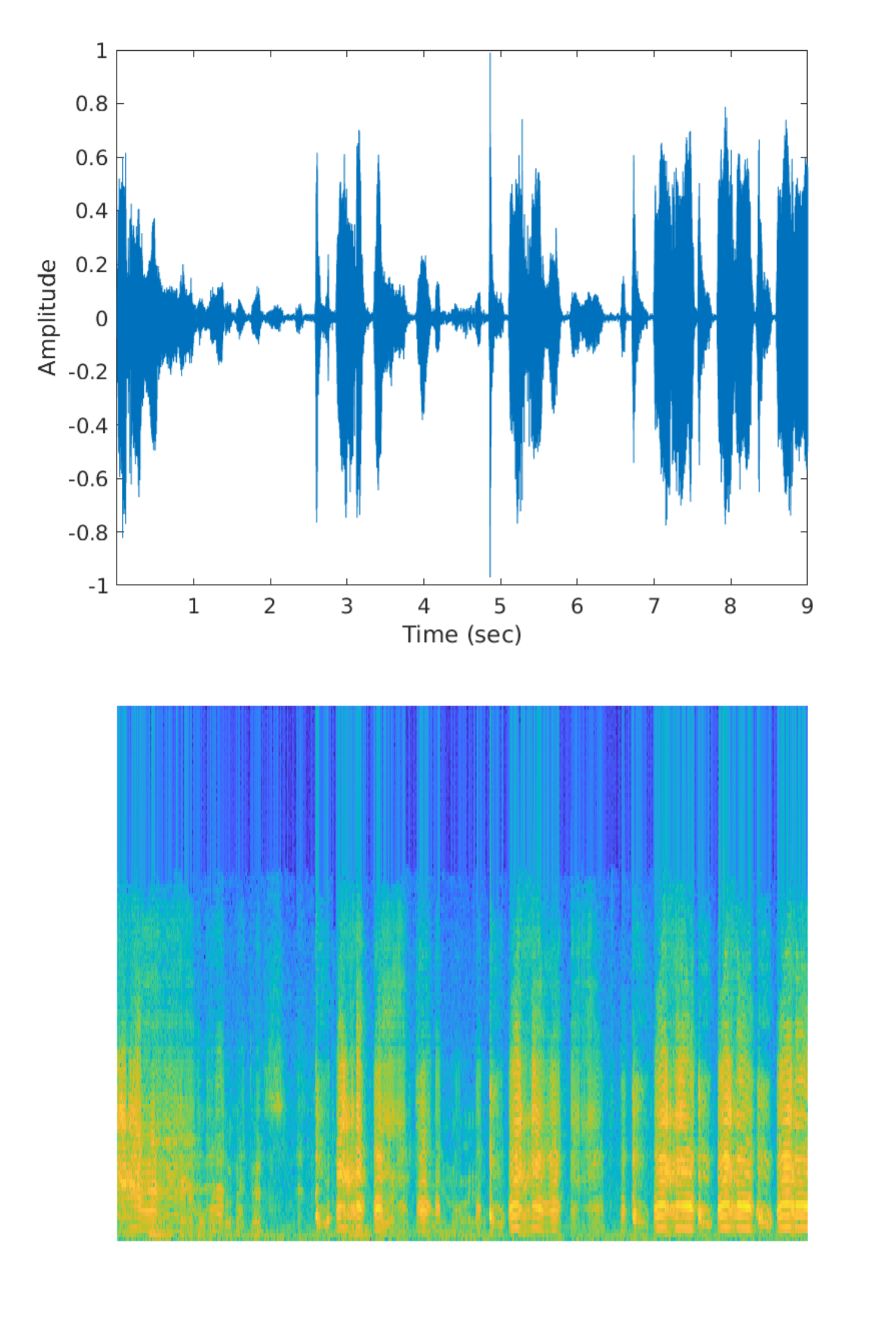}
\caption{Audio signal (top) and its corresponding spectrogram image (bottom) of a neonate during pain state.}
\label{Fig:spectrogram_pain}
\end{figure}

To train the network, we enlarged our set of spectrogram images by applying signal augmentation techniques to the original audio signal. Each audio signal is augmented by changing the raw frequency $f$ at 3 different levels ($f/3$, $f/2$, $2f/3$), and adding 6 different levels of noise ($0.001$, $0.003$, $0.005$, $0.01$, $0.03$, $0.05$). Further, a combination of both frequency and noise is also applied to create more variant signals. This process generated a total of 27 (3+6+3*6) augmented images for each audio signal. Fig. \ref{Fig:spectrogram_nopain} and Fig. \ref{Fig:spectrogram_pain} show examples of the raw audio signals and their corresponding spectrogram images during no-pain and pain states of a same subject.

\subsubsection{Feature Extraction}
Following the state-of-the-art methods \cite{zamzmi2019comprehensive, salekin2019harnessing}, we used both traditional machine learning classifier and deep learning-based classifier. In case of the traditional classifier, we trained classifiers such as Gaussian Naive Bayes \cite{rish2001empirical}, Random Forest \cite{breiman2001random}, and K-Nearest Neighbors (KNN) \cite{altman1992introduction} classifiers using the extracted MFCCs features.

For the deep learning-based classification, we used a pre-trained (ImageNet \cite{deng2009imagenet}) VGG-16 \cite{simonyan2014very} CNN network and fine-tuned this network (similar to Table \ref{Tab:vgg16_architecture}) using our postoperative pain dataset. The VGG-16 \cite{simonyan2014very} CNN network was trained using the spectrogram images extracted as described above. The last classification layer of the VGG-16 \cite{simonyan2014very} CNN has a $sigmoid$ activation function instead of the $linear$ activation.

\subsection{Multimodal Approach}
To generate a multimodal assessment of postoperative pain, we combined the pain scores generated by all indicator-specific models together using decision fusion as shown in Figure 4. The multimodal pain assessment is necessary because pain manifests itself in different signals \cite{zamzmi2018review, zamzmi2019comprehensive}. In addition, the multimodal approach is necessary because it allows to detect pain during the failure of recording some pain indicators as discussed in the next section and shown in Table \ref{Tab:unimodal}. To combine the labels or scores of facial expression, crying sound, and body movement, we used unweighted majority voting \cite{penrose1946elementary} scheme in which we choose the majority label in a given combination of labels as the final label. If the combination has a tie, we use the class probability (confidence score) to break the tie.

\begin{table*}
[width=.9\textwidth,cols=4,pos=h]
\caption{Unimodal and Multimodal assessment of neonatal postoperative pain using different traditional and deep learning approaches.}

\begin{tabular*}{\tblwidth}{@{} LLLLLLLLL@{} }

\toprule
Modality                & Approach                      &Accuracy &Precision &Recall &F1-Score &TPR &FPR &AUC \\ 
\midrule
\multirow{2}{*}{Face}   & VGG16 + LSTM                  &0.6203 &0.6195 &0.6203 &0.6197 &0.6634 &\textbf{0.4302} &0.7300 \\ \cline{2-9}
                        & \textbf{Bilinear VGG16 + LSTM}         &\textbf{0.6952} &\textbf{0.7084} & \textbf{0.6952} &\textbf{0.6834} &\textbf{0.8614} &0.5000 &\textbf{0.8196} \\ \cline{2-9}\hline
    
\multirow{5}{*}{Body}   & Motion + Gaussian NB          &0.6330 &0.6562 &0.6330 &0.6189 &0.4404 &\textbf{0.1743} &0.5001\\ \cline{2-9}
                        & Motion + Random Forest        &0.5872 &0.5874 &0.5872 &0.5868 &0.5596 &0.3853 &0.3382\\ \cline{2-9}
                        & Motion + KNN                  &0.5688 &0.5697 &0.5688 &0.5675 &0.5138 &0.3761 &0.3899\\ \cline{2-9}
                        & Motion Image + VGG16 + LSTM   &0.6835 &0.6906 &0.6835 &0.6805 &\textbf{0.7799} &0.4128 &0.7323\\ \cline{2-9}  
                        & \textbf{Body ROI Image + VGG16 + LSTM}                &\textbf{0.7050} &\textbf{0.7047} &\textbf{0.7050} &\textbf{0.7047} &0.7333 &0.3263 &\textbf{0.7786}\\ \cline{2-9}\hline

\multirow{4}{*}{Sound}  & MFCC + Gaussian NB            &0.6296 &0.6328 &0.6296 &0.6267 &0.5421 &0.2844 &0.4194\\ \cline{2-9}
                        & MFCC + KNN                    &0.6991 &0.7001 &0.6991 &0.6988 &0.7290 &0.3303 &0.3592 \\ \cline{2-9}
                        & MFCC + Random Forest          &0.7269	&0.7362	&0.7269	&0.7245	&\textbf{0.8224}	&0.3670	&0.4459 \\ \cline{2-9}
                        & \textbf{Spectrogram Image + VGG16}      &\textbf{0.7963}	&\textbf{0.7964}	&\textbf{0.7966}	&\textbf{0.7963} &0.7850	&\textbf{0.1927} &\textbf{0.8690} \\  \hline

Multimiodal & (\textbf{F+B+S}) + Decision Fusion & 0.7936 & \textbf{0.8028} & 0.7936 & \textbf{0.7920} & \textbf{0.8807} & 0.2936 & \textbf{0.9010} \\ 
\bottomrule                        

\multicolumn{9}{l}{* Precision, Recall, and F-1 score are weighted by both classes.} \\
\multicolumn{9}{l}{* TPR, FPR, and AUC are calculated for the pain class.} \\
\multicolumn{9}{l}{* Bold texts indicate our approaches and bold values indicate superiority.} \\
\multicolumn{9}{l}{* Bold text (\textbf{F+B+S}) represents the best from the unimodal (bold texts) approaches.} \\
\end{tabular*}
\label{Tab:unimodal}
\end{table*}

\begin{table*}
[width=.9\textwidth,cols=4,pos=h]
\caption{Unimodal and Multimodal neonatal assessment of postoperative pain (all pain indicators are present).}
  \begin{tabular*}{\tblwidth}{@{} LLLLLLLL@{} }
\toprule
    Metric          &Face   &Body   &Sound  &Face + Body   &Body + Sound  &Sound + Face  &Face + Body + Sound  \\
    \midrule
    Accuracy        &0.7076	&0.6667	&0.7661	&0.7076 &0.7719 &0.6901 &\textbf{0.7895} \\ 
    Precision       &0.7119	&0.6645	&0.7682	&0.8071 &\textbf{0.8274} &0.7032 &0.7913 \\ 
    Recall          &0.7076	&0.6667	&0.7661	&0.7076 &0.7719 &0.6901 &\textbf{0.7895} \\ 
    F-1 Score       &0.6970	&0.6650	&0.7667	&0.6630 &0.7522 &0.6703 &\textbf{0.7863} \\ 
    TPR             &0.8557	&0.7320	&0.7732	&\textbf{1.0000} &0.9897 &0.8866 &0.8761\\ 
    FPR             &0.4865	&0.4189	&\textbf{0.2432}	&0.6757 &0.5135 &0.5676 &0.3243 \\
    AUC             &0.8082	&0.7778	&0.8239	&0.8353 &0.8763 &0.8396 &\textbf{0.8791} \\ 
    \bottomrule
\multicolumn{8}{l}{* Precision, Recall, and F-1 score are weighted by both classes.} \\
\multicolumn{8}{l}{* TPR, FPR, and AUC are calculated for the pain class.} \\
\multicolumn{8}{l}{* Bold values indicate superiority.} \\

\end{tabular*}
\label{Tab:modal_innerjoin}
\end{table*}

\begin{table*}
[width=.9\textwidth,cols=4,pos=h]
\caption{Unimodal and Multimodal assessment of neonatal postoperative pain (randomly dropping 25\% samples from each indicator 10 times).}
\begin{tabular*}{\tblwidth}{@{} LLLLLLL@{} }
\toprule
    \multirow{2}{*}{Metric} 
    &\multicolumn{2}{C}{Face} &\multicolumn{2}{C}{Body}  &\multicolumn{2}{C}{Sound} \\ 
    \cline{2-3} \cline{4-5} \cline{6-7}
    &Unimodal         &Multimodal           &Unimodal         &Multimodal           &Unimodal          &Multimodal  \\
    \midrule
    Accuracy                &0.7124 $\pm$ 0.03 &\textbf{0.7913$\pm$ 0.01} 	&0.6610$\pm$ 0.02 &\textbf{0.7649$\pm$ 0.01}	    &0.7742 $\pm$ 0.01 &\textbf{0.7784 $\pm$ 0.01} 	\\ 
    Precision               &0.7218 $\pm$ 0.03 &\textbf{0.7988$\pm$ 0.01}	&0.6596$\pm$ 0.02 &\textbf{0.7692$\pm$ 0.01}	    &0.7764 $\pm$ 0.01 &\textbf{0.7908 $\pm$ 0.01}	\\ 
    Recall                  &0.7124 $\pm$ 0.03 &\textbf{0.7913$\pm$ 0.01}	&0.6610$\pm$ 0.02 &\textbf{0.7650$\pm$ 0.01}	    &0.7742 $\pm$ 0.01 &\textbf{0.7784 $\pm$ 0.01}	\\ 
    F-1 Score               &0.7035 $\pm$ 0.03 &\textbf{0.7859$\pm$ 0.01}	&0.6591$\pm$ 0.02 &\textbf{0.7593$\pm$ 0.01}	    &\textbf{0.7746 $\pm$ 0.01} &0.7705 $\pm$ 0.01	\\ 
    TPR                     &0.8563 $\pm$ 0.03 &\textbf{0.9052$\pm$ 0.02}	&0.7282$\pm$ 0.03 &\textbf{0.8784$\pm$ 0.00}	    &0.7819 $\pm$ 0.03 &\textbf{0.9155 $\pm$ 0.02	}\\ 
    FPR                     &0.4612 $\pm$ 0.04 &\textbf{0.3581$\pm$ 0.03}	&0.4250$\pm$ 0.03 &\textbf{0.3838$\pm$ 0.02}	    &\textbf{0.2358 $\pm$ 0.03} &0.4014 $\pm$ 0.03   \\ 
    AUC                     &0.8093 $\pm$ 0.02 &\textbf{0.8724$\pm$ 0.01}	&0.7739$\pm$ 0.02 &\textbf{0.8675$\pm$ 0.01}     &0.8288 $\pm$ 0.02 &\textbf{0.8682 $\pm$ 0.01}   \\ 
\bottomrule
\multicolumn{7}{l}{* Precision, Recall, and F-1 score are weighted by both classes.} \\
\multicolumn{7}{l}{* TPR, FPR, and AUC are calculated for the pain class.} \\
\multicolumn{7}{l}{* Bold values indicate superiority.} \\

\end{tabular*}
\label{Tab:multimodal_robustess}
\end{table*}

\section{Experimental Results and Discussion}
\label{Sec:results_discussion}
In this section, we present the performance of assessing neonatal postoperative pain using a single pain indicator at a time (unimodal) and multiple pain indicators together (multimodal). Before presenting the results, we describe the process of extracting and preparing the videos followed by our training and evaluation protocols.

\subsection{Dataset Preparation}
We used the aforementioned (Section \ref{Sec:neonatal_pain_dataset}) neonatal pain dataset to evaluate the proposed temporal multimodal approach. The dataset consists of both procedural (202 videos) and postoperative (218 videos) pain. We used procedural dataset  a balanced set of 116 samples) for pre-training the model (in case of face only), and used the postoperative dataset for fine-tuning and evaluation. After performing the pre-processing steps (see Section \ref{Sec:methodology}), the total number of video segments (each has 9 seconds length) for each pain indicator in the postoperative dataset, were 187, 218, and 216 for face, body, and sound, respectively. Note that the face was missing in 31 videos (187/218) and the sound was missing in 2 videos (216/218).

\subsection{Training and Evaluation Protocol}
We used two types of training techniques: traditional classifiers training and deep learning. For both cases, we used the leave-one-subject-out protocol for training and testing as this protocol is more realistic in case of clinical applications (see Ref. \cite{saeb2017need, koul2018cross}) as it allows to capture the differences between patients. In the case of the traditional classifiers, we used KNN \cite{altman1992introduction} classifier ($K=3$, determined empirically), and Random Forest \cite{breiman2001random} classifier ($N=100$, determined empirically). For deep learning, we used images (face image, body image, motion image, and spectrogram) of size $224 \times 224$ as input to individual VGG-16 \cite{simonyan2014very} models to extract deep features from each individual indicator as shown in Fig. \ref{Fig:multimodal_approach}. The extracted features are then fed to RNN networks to learn pain patterns and dynamics. We used Adam \cite{kingma2014adam} optimizer with a learning rate of 0.0001 to train the CNN and RNN models. A batch size of 16 and 1 are used for CNN and RNN respectively for up to 100 epochs. All the training is performed to minimize the validation loss following an early stopping strategy.

We performed two levels of training in case of deep learning. In the first level, we used the pain scores of each indicator (i.e., score 0 or 1 [face and body] and score 0, 1, or 2 [sound]) for training the CNN models. In the second level, we used the final pain labels, which are no-pain, moderate pain, and severe pain, to train the RNN models. As discussed in Section \ref{Sec:neonatal_pain_dataset}, these final pain labels are generated by summing the individual scores and thresholding. Note that we combined the labels of moderate and severe pain into a single pain class while training the RNN models because the number of instances with a moderate pain label is relatively smaller (33 examples).

To evaluate the performance of the trained models, we used the weighted accuracy, weighted precision, weighted recall, and F-1 score. Weighted metrics reflect the performance of each class as they report the fraction of the correct prediction for each class over the total number of samples; i.e., weighted metrics consider the instances of a specific class. In addition to these, we calculated the True Positive Rate (TPR), False Positive Rate (FPR), and Area Under the Curve (AUC) for the pain class.

\subsection{Unimodal Postoperative Pain Assessment}
We evaluated the performance of using a single pain indicator at a time for postoperative pain assessment. We used both traditional machine learning-based approaches and deep learning-based approaches. Table \ref{Tab:unimodal} shows the performance of using both traditional and deep learning approaches with a single pain indicator for assessing postoperative pain. In all indicators and in most cases, our approaches outperformed the state-of-the-art methods \cite{zamzmi2019comprehensive} by a large margin. As can be seen from Table \ref{Tab:unimodal}, crying sound indicator achieved the highest accuracy (79.63\%) and outperformed the accuracies of body (70.50\%) and face (69.52\%). Similarly, crying sound indicator achieved the highest AUC (0.87) and outperformed the AUCs of body (0.78) and face (0.82).

To understand these results, we observed the data and found that sound has less noise as compared to face and body in our dataset of postoperative neonates. Specifically, neonates' faces in the NICU are usually occluded (partial or complete) by oxygen's masks, tapes, or due to a prone sleeping position. In case of body, some neonates are swaddled while others show weak movements due to sedation or exhaustion. In summary, we can conclude from the Table \ref{Tab:modal_innerjoin} that crying sound can better assess postoperative pain as compared to facial expression and body movement. In addition, we can conclude that our proposed approaches for analyzing facial expression, sound, and body show better performance, in terms of accuracy, precision, recall, TPR, FPR, and AUC, as compared to the traditional approaches.

In addition, it can also be observed that temporal information integration improves the performance a lot. Existing works \cite{zamzmi2019comprehensive, zamzmi2019convolutional}, did consider the feature only frame-by-frame. But we integrate temporal information (over frames) which leads the better performance in case of all approaches. In case of body, inclusion of the LSTM network shows AUC of 0.78 and 0.73 which was a jump from 0.50. Also, in case of sound, the spectrogram image shows better performance compared to the MFCC features due to better temporal information integration.

\subsection{Multimodal Postoperative Pain Assessment}
\begin{figure}
\centering
\includegraphics[width=.48\textwidth]{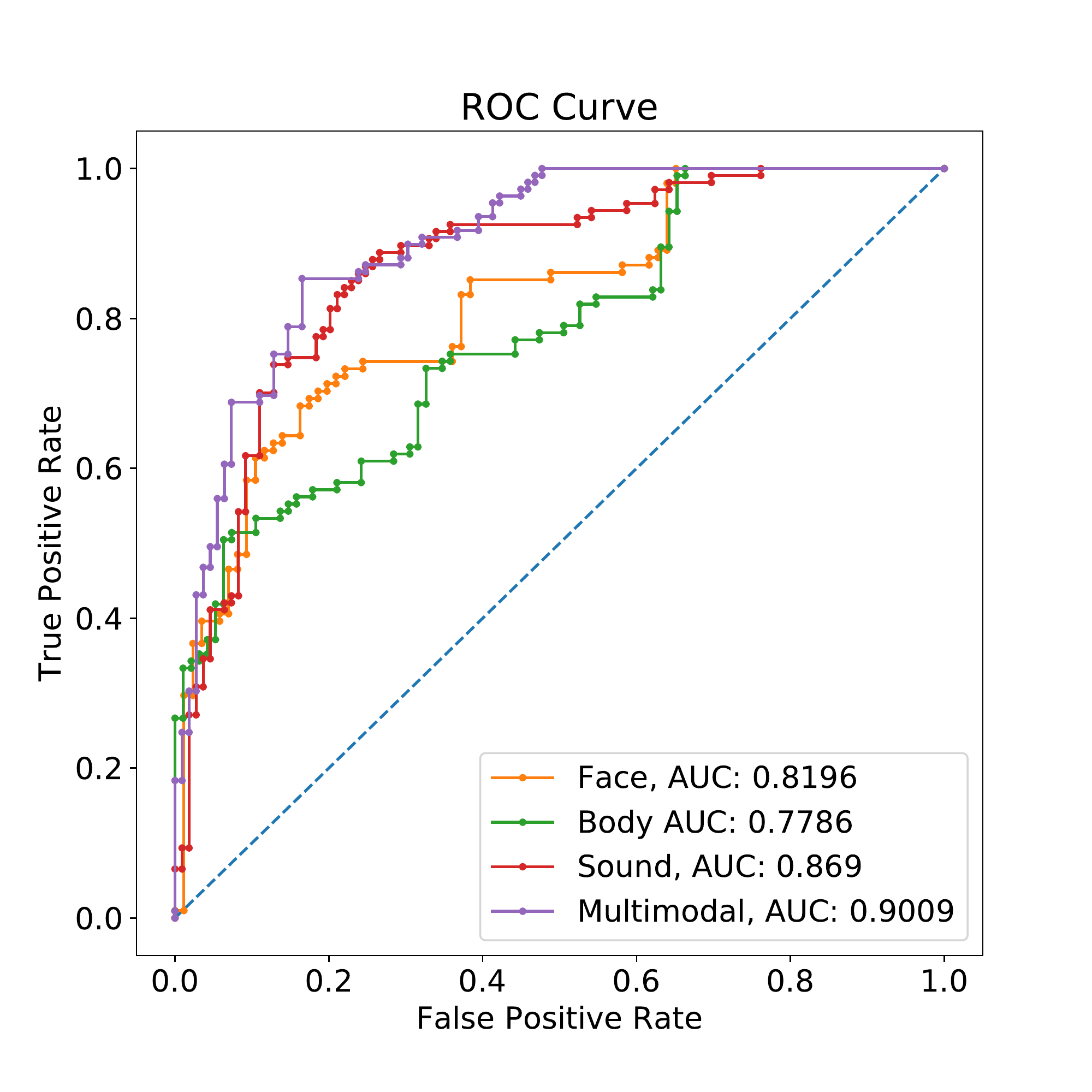}
\caption{ROC curves of different approaches.}
\label{Fig:roc_curve}
\end{figure}

The unimodal approach uses a single indicator at a time to predict the pain class. In practice, there are cases where face and body are not visible. For example, the baby's face can be wrapped with tape and the body can be swaddled. In such cases, the multimodal assessment provides a reliable solution \cite{temko2015multimodal}. To investigate the impact of the multimodal approach on postoperative pain assessment, we combined the scores or labels of different pain indicators, which are generated using the best approach for each indicator (best approaches are bolded in the second column of Table \ref{Tab:unimodal}). Table \ref{Tab:unimodal} shows the results of fusing (decision-level) the labels of face, body, and sound. Recall that the numbers of video instances for face, body, and sound are 187, 218, and 216, respectively. This means that some indicators would be missing when we combine all of them together to generate the multimodal assessment. As shown in Table \ref{Tab:unimodal}, the multimodal approach achieved better overall performance as compared to the unimodal approach. The reason for the high performance of sound can be attributed to the fact that this indicator has less noise and a larger number of instances as compared to other indicators (e.g., facial expression). Although crying sound has a similar performance compared to the multimodal approach, we believe that the multimodal approach is necessary because pain manifests itself in different signals. In addition, the multimodal approach allows to assess pain during circumstances when sounds signals are missing due to noise, sedation, or individual differences (e.g., some neonates do not cry but move their arms/legs during pain). Fig. \ref{Fig:roc_curve} provides visualization of the ROC curve of Table \ref{Tab:unimodal}. It can be observed that the multimodal approach achieves better performance (curve) compared to the individual modalities.

To make a more reliable and fair comparison, we further extend our experiments by making sure that there are no missing indicators; i.e., we selected 171 samples from our dataset where all the pain indicators are present. Table \ref{Tab:modal_innerjoin} presents the performance of the multimodal when all indicators are present. The Table also presents the performance of unimodal (single indicator at a time) and different combinations of pain indicators using 171 samples. It can be observed that in most cases the multimodal achieved the best performance. In the final experiment, we randomly dropped 25\% of samples from each indicator to assess the robustness of our multimodal approach. We performed random dropping by 25\% ten times and reported the average performance in Table \ref{Tab:multimodal_robustess}. From the Table \ref{Tab:multimodal_robustess}, we can conclude that the multimodal results are consistent over all indicators and perform better than the unimodal. These results are consistent with previous clinical findings \cite{hudson2002validation,hummel2008clinical} and suggest that the automated multimodal approach for assessing postoperative pain is more efficient, in terms of performance and robustness, as compared to the unimodal approach.

\section{Conclusion and Future Work}
\label{Sec:conclusion}
In this paper, a temporal multimodal AI-based system is proposed for assessing postoperative pain in neonates. The proposed system uses video (face, body) and audio (crying sound) signals individually to generate pain scores. These scores are then combined using a decision fusion to predict the final pain assessment. We compared the proposed multimodal approach with the traditional machine learning approaches and found that our approach achieved superior performance. We also found that the multimodal approach is better than the unimodal approach for assessing postoperative in neonates. The experimental results suggest that the multimodal approach is more reliable for assessing postoperative pain in a real-world clinical environment. We believe that the proposed approach can significantly enhance the current assessment practice, which is discontinuous, inconsistent, highly depends on the nurses' experience and subjectivity, and is often limited by the lack of medical resources. In the future, we plan to integrate other signals, such as vital signs, into our multimodal system. We also plan to investigate other fusion methods such as feature level fusion.

%%%%%%%%%%%%%%%%%%%%%%%%%%%%%%%%%%%%%%%%%%%%%%%%%%%%%%%%%%%%%%%%%%%%%%%%%%%%%%%%

\section*{Competing interests}
The authors declare no competing interests.

\section*{Acknowledgment}
We are grateful for the entire neonatal staff at Tampa General Hospital for their help and cooperation in the data collection. 
% This research is partially supported by University of South Florida Nexus Initiative (UNI) Grant and National Institutes of Health (NIH), United States Grant (NIH R21NR018756).
This research is partially supported by University of South Florida Nexus Initiative (UNI) Grant and National Institutes of Health Grant (NIH R21NR018756).

% supported by National Institute of Nursing Research (NINR) of the National Institutes of Health under award number R21NR018756.

% supported by the National Institutes of Health under award number R21NR018756.

%%%%%%%%%%%%%%%%%%%%%%%%%%%%%%%%%%%%%%%%%%%%%%%%%%%%%%%%%%%%%%%%%%%%%%%%%%%%%%%%

\bibliographystyle{cas-model2-names}
\bibliography{ref}

\end{document}